\documentclass[a4paper]{article}

\usepackage{INTERSPEECH2016}

\usepackage{graphicx}
\usepackage{amssymb,amsmath,bm}
\usepackage{textcomp}
\usepackage{cite}
\usepackage{multirow}
\usepackage{float}
\usepackage{hyperref}
\ninept

\title{Towards Machine Comprehension of Spoken Content:\\
Initial TOEFL Listening Comprehension Test by Machine
}

\makeatletter
\def\name#1{\gdef\@name{#1\\}}
\makeatother \name{{\em Bo-Hsiang Tseng, Sheng-Syun Shen, Hung-Yi Lee, Lin-Shan Lee}}

\address{Graduate Institute of Communication Engineering\\ National Taiwan University \\
{\footnotesize \tt r02942037@ntu.edu.tw, r03942071@ntu.edu.tw, tlkagkb93901106@gmail.com, lslee@gate.sinica.edu.tw}
\thanks{This work was supported by Ministry and Science Technology R.O.C and MediaTek Inc. under Contract MOST 104-2622-8-002-002}
}

\usepackage{subfig}
\begin{document}

  \maketitle
  \begin{abstract}

Multimedia or spoken content presents more attractive information than plain text content, but it's more difficult to display on a screen and be selected by a user.
As a result, accessing large collections of the former is much more difficult and time-consuming than the latter for humans.
It's highly attractive to develop a machine which can automatically understand spoken content and summarize the key information for humans to browse over.
In this endeavor, we propose a new task of machine comprehension of spoken content.
We define the initial goal as the listening comprehension test of TOEFL, a challenging academic English examination for English learners whose native language is not English.
We further propose an Attention-based Multi-hop Recurrent Neural Network (AMRNN) architecture for this task, achieving encouraging results in the initial tests.
Initial results also have shown that word-level attention is probably more robust than sentence-level attention for this task with ASR errors.

  \end{abstract}

  \section{Introduction} \label{sec:intro}
  With the popularity of shared videos, social networks, online course, etc, the quantity of multimedia or spoken content is growing much faster beyond what human beings can view or listen to.
Accessing large collections of multimedia or spoken content is difficult and time-consuming for humans, even if these materials are more attractive for humans than plain text information.
Hence, it will be great if the machine can automatically listen to and understand the spoken content, and even visualize the key information for humans.
This paper presents an initial attempt towards the above goal: machine comprehension of spoken content.
In an initial task, we wish the machine can listen to and understand an audio story, and answer the questions related to that audio content.
TOEFL listening comprehension test is for human English learners whose native language is not English.
This paper reports how today's machine can perform with such a test.

The listening comprehension task considered here is highly related to Spoken Question Answering (SQA)~\cite{i2012factoid,ispoken}.
In SQA, when the users enter questions in either text or spoken form, the machine needs to find the answer from some audio files.
SQA usually worked with ASR transcripts of the spoken content, and used information retrieval (IR) techniques~\cite{shiang2014spoken} or relied on knowledge bases~\cite{hixon2015learning} to find the proper answer.
Sibyl~\cite{sibyl}, a factoid SQA system, used some IR techniques and utilized several levels of linguistic information to deal with the task.
Question Answering in Speech Transcripts (QAST)~\cite{QAST09,QAST08,QAST07} has been a well-known evaluation program of SQA for years.
However, most previous works on SQA mainly focused on factoid questions like \textit{``What is name of the highest mountain in Taiwan?''}.
Sometimes this kind of questions may be correctly answered by simply extracting the key terms from a properly chosen utterance without understanding the given spoken content.
More difficult questions that cannot be answered without understanding the whole spoken content seemed rarely dealt with previously.

With the fast development of deep learning, neural networks have successfully applied to speech recognition\cite{dahl2013improving,deng2013new,graves2013speech} or NLP tasks\cite{kalchbrenner2014convolutional,collobert2011natural}.
A number of recent efforts have explored various ways to understand multimedia in text form\cite{weston2014memory,hermann2015teaching,bordes2015large,sukhbaatar2015end,kumar2015ask,rush2015neural}.
They incorporated attention mechanisms\cite{sukhbaatar2015end} with Long Short-Term Memory based networks\cite{hochreiter1997long}.
In Question Answering field, most of the works focused on understanding text documents\cite{bordes2014question,er2013factoid,iyyer2014neural,fader2014open}.
Even though ~\cite{tapaswi2015movieqa} tried to answer the question related to the movie, they only used the text and image in the movie for that.
It seems that none of them have studied and focused on comprehension of spoken content yet.

           \begin{figure}[!bt]
        \centering
        \includegraphics[width=1\linewidth]{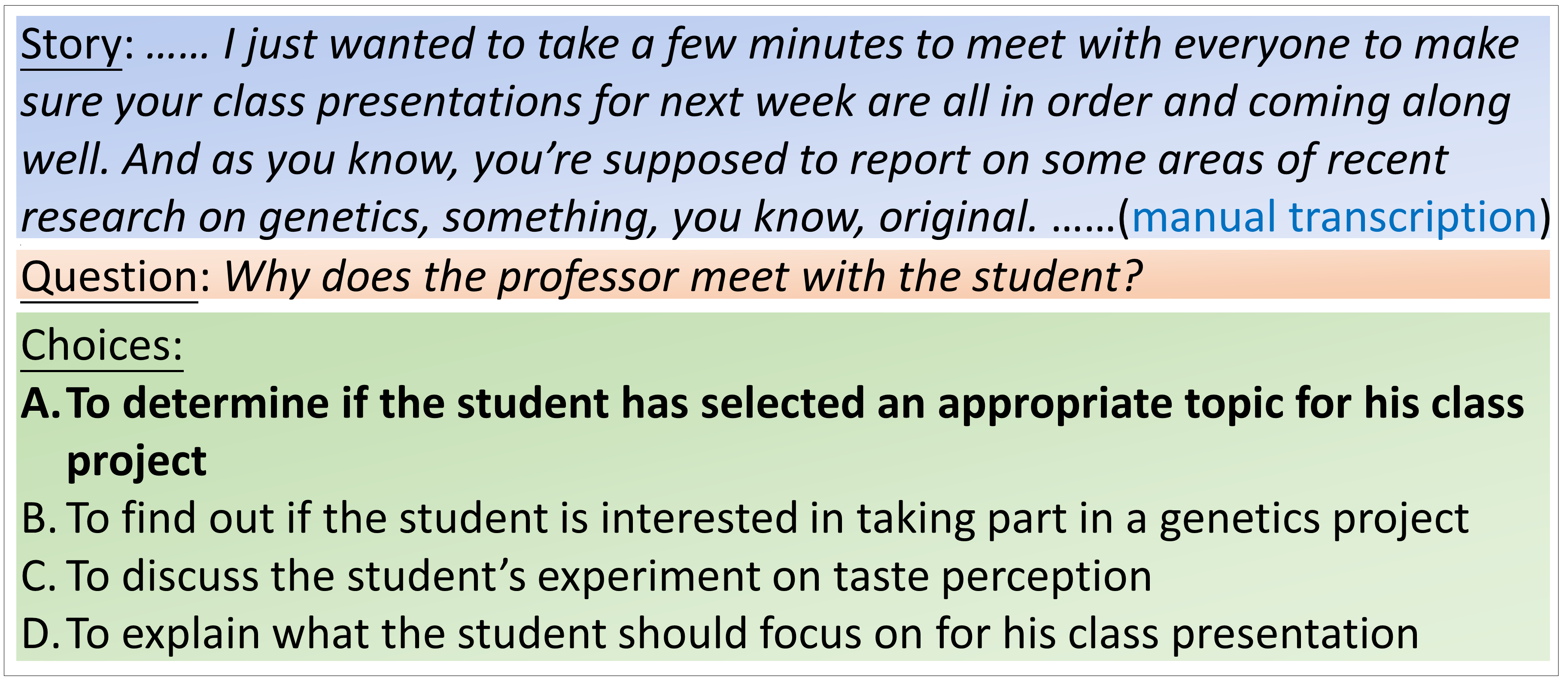}
		  \caption{An example of TOEFL listening comprehension test. 
          The story is given in audio format, and its manual transcription is shown. The question and choices are provided in text format.
          }
        \label{fig:example}
      \end{figure}

  \section{Task Definition and Contributions} \label{sec:task}
In this paper, we develop and propose a new task of machine comprehension of spoken content which was never mentioned before to our knowledge.
We take TOEFL listening comprehension test as an corpus for this work.
TOEFL is an English examination which tests the knowledge and skills of academic English for English learners whose native languages is not English.
In this examination, the subjects would first listen to an audio story around five minutes and then answer several question according to that story.
The story is related to the college life such as conversation between the student and the professor or a lecture in the class. 
Each question has four choices where only one is correct.
An real example in the TOEFL examination is shown in Fig.~\ref{fig:example}.
The upper part is the manual transcription of a small part of the audio story.
The questions and four choices are listed too.
The correct choice to the question in Fig.~\ref{fig:example} is choice A. 
The questions in TOEFL are not simple even for a human with relatively good knowledge because the question cannot be answered by simply matching the words in the question and in the choices with those in the story, and key information is usually buried by many irrelevant utterances. 
To answer the questions like \textit{``Why does the student go to professor's office?"}, the listeners have to understand the whole audio story and draw the inferences to answer the question correctly.
As a result, this task is believed to be very challenging for the state-of-the-art spoken language understanding technologies.



We propose a listening comprehension model for the task defined above, the Attention-based Multi-hop Recurrent Neural Network (AMRNN) framework, and show that this model is able to perform reasonably well for the task. 
In the proposed approach, the audio of the stories is first transcribed into text by ASR, and the proposed model is developed to process the transcriptions for selecting the correct answer out of 4 choices given the question. 
The initial experiments showed that the proposed model achieves encouraging scores on the TOEFL listening comprehension test.
The attention-mechanism proposed in this paper can be applied on either word or sentence levels. 
We found that sentence-level attention achieved better results on the manual transcriptions without ASR errors, but word-level attention outperformed the sentence-level on ASR transcriptions with errors. 

        \begin{figure}[tb]
        \centering
        \includegraphics[width=1\linewidth]{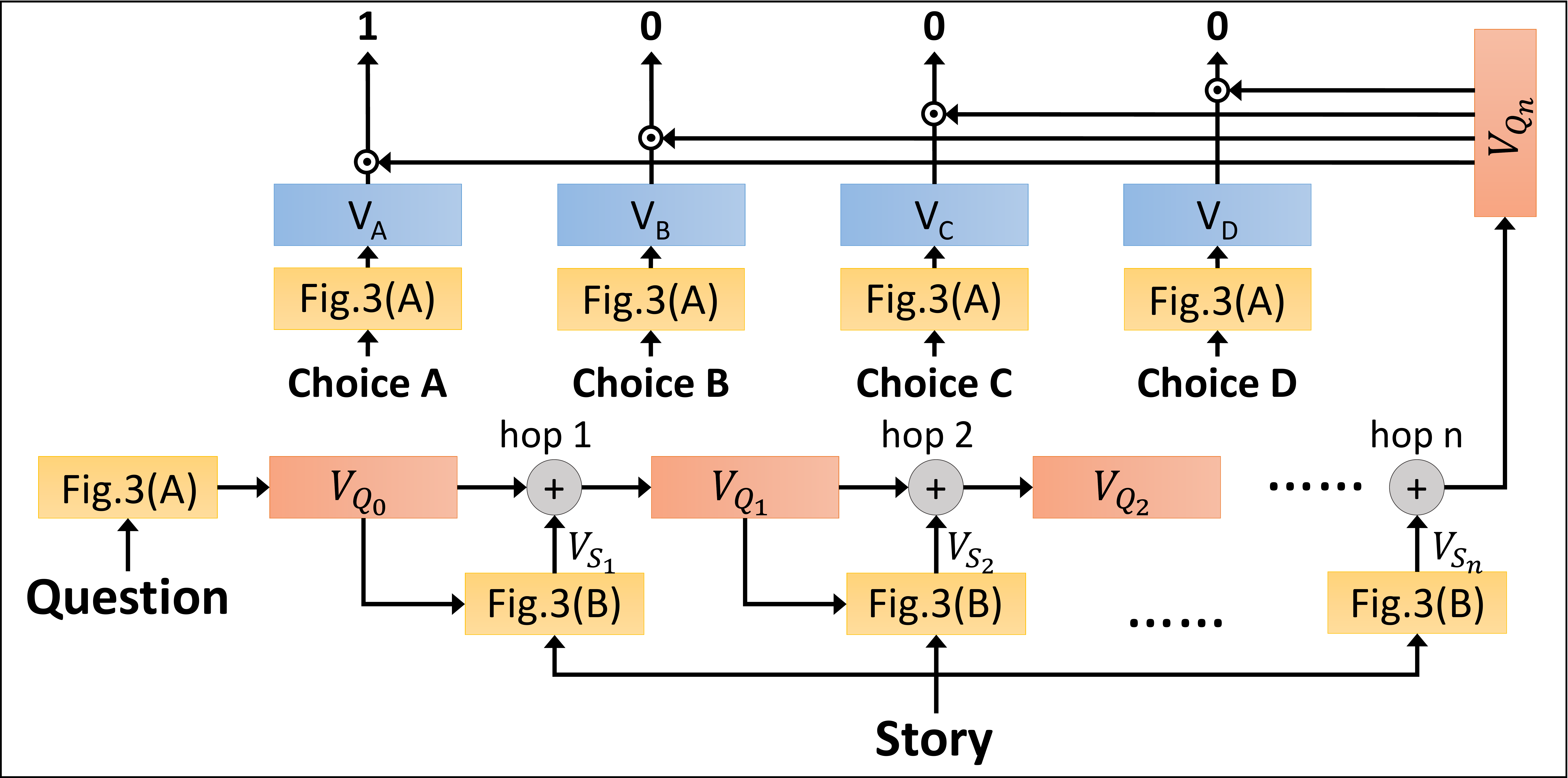}
		  \caption{The overall structure of the proposed Attention-based Multi-hop Recurrent Neural Network (AMRNN) model.}
        \label{fig:overview}
      \end{figure}
  
	\begin{figure}[tb]
        \centering
        \includegraphics[width=1\linewidth]{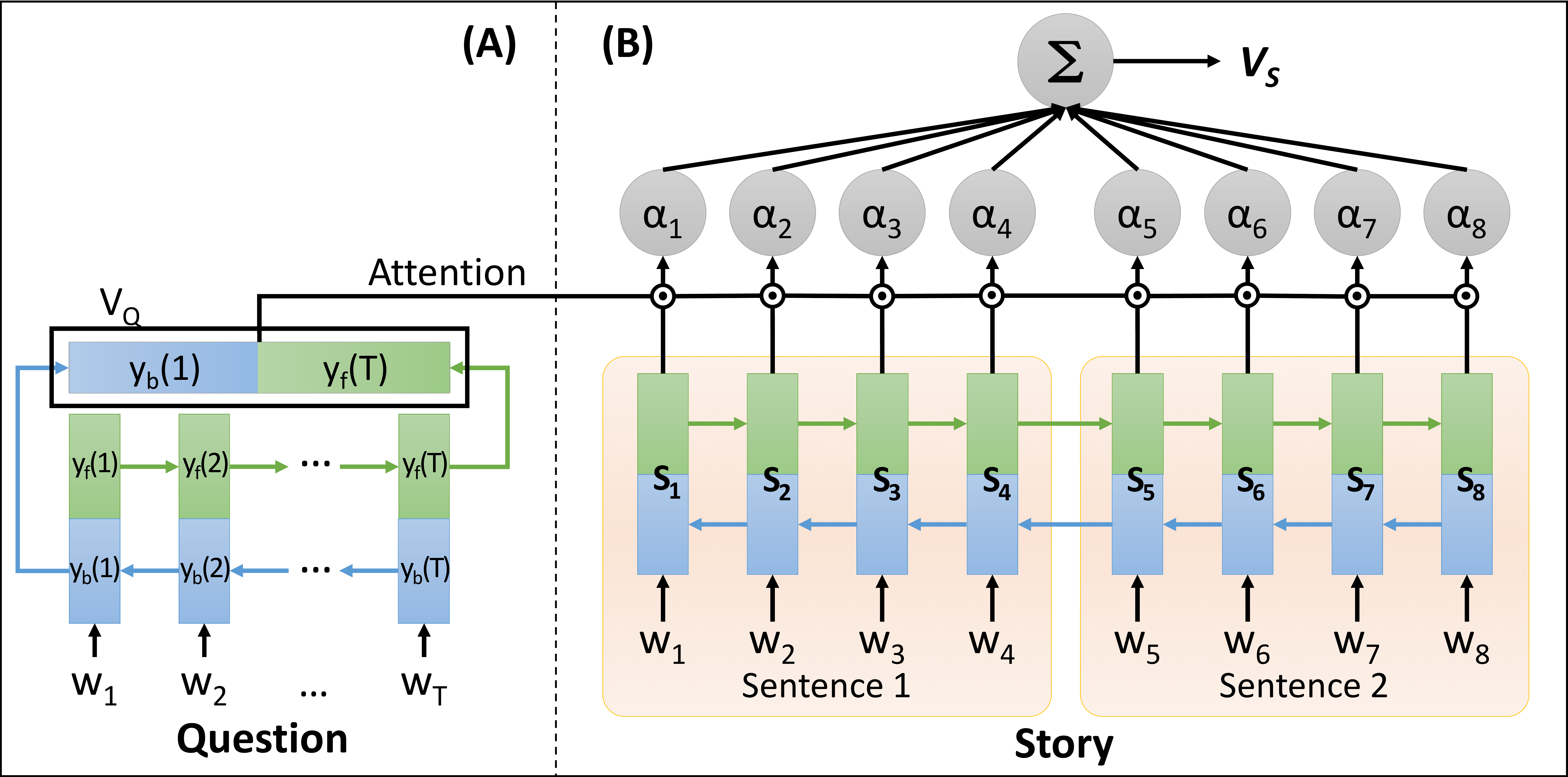}
		  \caption{(A) The Question Vector Representation and (B) The Attention Mechanism.}
        \label{fig:attention}
      \end{figure}
   
\section{Proposed Approach} \label{sec:propose}
The overall structure of the proposed model is in Fig~\ref{fig:overview}.
The input of model includes the transcriptions of an audio story, a question and four answer choices, all represented as word sequences.
The word sequence of the input question is first represented as a question vector $V_Q$ in Section~\ref{subsec:question}.
With the question vector $V_Q$, the attention mechanism is applied to extract the question-related information from the story in Section~\ref{sec:attention}.
The machine then goes through the story by the attention mechanism several times and obtain an answer selection vector $V_{Q_n}$ in Section~\ref{sec:hop}.
This answer selection vector $V_{Q_n}$ is finally used to evaluate the confidence of each choice in Section~\ref{subsec:select}, and the choice with the highest score is taken as the output.
All the model parameters in the above procedure are jointly trained with the target where 1 for the correct choice and 0 otherwise.

\subsection{Question Representation} \label{subsec:question}
Fig.~\ref{fig:attention} (A) shows the procedure of encoding the input question into a vector representation $V_Q$.
The input question is a sequence of T words, {$w_1,w_2,...,w_T$}, every word $W_{i}$ represented in 1-Of-N encoding.
A bidirectional Gated Recurrent Unit (GRU) network\cite{chung2014empirical,cho2014properties,bahdanau2014neural} takes one word from the input question sequentially at a time. 
In Fig~\ref{fig:attention} (A), the hidden layer output of the forward GRU (green rectangle) at time index $t$ is denoted by $y_{f}(t)$, and that of the backward GRU (blue rectangle) is by $y_{b}(t)$. 
After looking through all the words in the question, the hidden layer output of forward GRU network at the last time index $y_{f}(T)$, and that of backward GRU network at the first time index $y_{b}(1)$, are concatenated to form the question vector representation $V_{Q}$, or $V_{Q} = [y_{f}(T) \| y_{b}(1)]$\footnote{The symbol [$\cdot$$\|$$\cdot$] denotes concatenation of two vectors in this paper.}. 

\subsection{Story Attention Module} \label{sec:attention}
Fig.~\ref{fig:attention} (B) shows the attention mechanism which takes the question vector $V_Q$ obtained in Fig.~\ref{fig:attention} (A) and the story transcriptions as the input to encode the whole story into a story vector representation $V_{S}$.
The story transcription is a very long word sequence with many sentences, so we only show two sentences each with 4 words for simplicity.
There is a bidirectional GRU in Fig~\ref{fig:attention} (B) encoding the whole story into a story vector representation $V_{S}$. 
The word vector representation of the $t$-th word $S_{t}$ is constructed by concatenating the hidden layer outputs of forward and backward GRU networks, that is $S_t = [y_{f}(t) \| y_{b}(t)]$.
Then the attention value $\alpha_t$ for each time index ${t}$ is the cosine similarity between the question vector $V_{Q}$ and the word vector representation $S_{t}$ of each word, $\alpha_t = S_t \odot V_{Q}$\footnote{The symbol $\odot$ denotes cosine similarity between two vectors.}. 
With attention values $\alpha_t$, there can be two different attention mechanisms, word-level and sentence-level, to encode the whole story into the story vector representations $V_{S}$.

\textbf{Word-level Attention}: We normalize all the attention values $\alpha_t$ into $\alpha_t^\prime$ such that they sum to one over the whole story.
Then all the word vector $S_{t}$ from the bidirectional GRU network for every word in the story are weighted with this normalized attention value $\alpha_{t}^\prime$ and sum to give the story vector, that is $V_{S} = \sum_{t}\alpha_{t}^{\prime}S_{t}$. 

\textbf{Sentence-level Attention}: 
Sentence-level attention means the model collects the information only at the end of each sentence. 
Therefore, the normalization is only performed over those words at the end of the sentences to obtain $\alpha_t^{\prime\prime}$.
The story vector representation is then $V_{S} = \sum_{t=eos}\alpha_t^{\prime\prime}*S_{t}$, where only those words at the end of sentences (eos) contribute to the weighted sum. 
So $V_{S} = \alpha_4^{\prime\prime}*S_4 + \alpha_8^{\prime\prime}*S_8$ in the example of the Fig.\ref{fig:attention}
\subsection{Hopping} \label{sec:hop} 
The overall picture of the proposed model is shown in Fig~\ref{fig:overview}, in which Fig.~\ref{fig:attention} (A) and (B) are component modules (labeled as Fig.~\ref{fig:attention} (A) and (B)) of the complete proposed model.
In the left of Fig.~\ref{fig:overview}, the input question is first converted into a question vector $V_{Q_0}$ by the module in Fig.~\ref{fig:attention} (A). 
This $V_{Q_0}$ is used to compute the attention values {$\alpha_{t}$} to obtain the story vector $V_{S_1}$ by the module in Fig.~\ref{fig:attention} (B).
Then $V_{Q_0}$ and $V_{S_1}$ are summed to form a new question vector $V_{Q_1}$.
This process is called the first hop (hop 1) in Fig.~\ref{fig:overview}.
The output of the first hop $V_{Q_1}$ can be used to compute the new attention to obtain a new story vector $V_{S_1}$.
This can be considered as the machine going over the story again to re-focus the story with a new question vector.
Again, $V_{Q_1}$ and $V_{S_1}$ are summed to form $V_{Q_2}$ (hop 2).
After $n$ hops ($n$ should be pre-defined), the output of the last hop $V_{Q_n}$ is used for the answer selection in the Section~\ref{subsec:select}.

   	\subsection{Answer Selection} \label{subsec:select}
    As in the upper part of Fig.~\ref{fig:overview}, the same way previously used to encode the question into $V_Q$ in Fig.~\ref{fig:attention} (A) is used here to encode four choice into choice vector representations $V_A$, $V_B$, $V_C$, $V_D$.
Then the cosine similarity between the output of the last hop $V_{Q_n}$ and the choice vectors are computed, and the choice with highest similarity is chosen.
	\begin{figure*}
        \centering
        \includegraphics[width=0.99\linewidth]{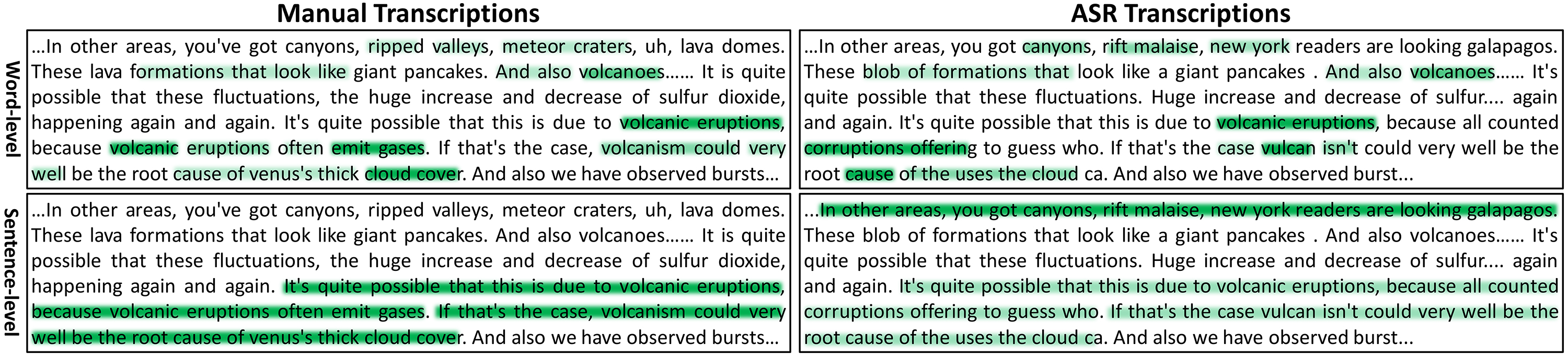}
		  \caption{{Visualization of the attention weights in sentence-level and in word-level on a small section of the manual or ASR transcriptions of an example story given a question.
          The darker the color, the higher the weights.
          The question of this story is \textit{``What is a possible origin of Venus'clouds?"} and the correct answer choice is \textit{``Gases released as a result of volcanic activity"}.}}
        \label{fig:attention_example}
      \end{figure*}

  \section{Experiments} \label{sec:exp}
  
  \subsection{Experimental Setup} \label{sec:setup}
    
\hspace*{\parindent}
	$\bullet$ Dataset Collection:  The collected TOEFL dataset included 963 examples in total (717 for training, 124 for validation, 122 for testing).
  Each example included a story, a question and 4 choices. 
  Besides the audio recording of each story, the manual transcriptions of the story are also available.
  We used a pydub library~\cite{pydub} to segment the full audio recording into utterances.
  Each audio recording has 57.9 utterances in average. 
  There are in average 657.7 words in a story, 12.01 words in question and 10.35 words in each choice.

	$\bullet$ Speech Recognition:   We used the CMU speech recognizer - Sphinx~\cite{walker2004sphinx} to transcribe the audio story.
  The recognition word error rate (WER) was 34.32\%.
  
  
  $\bullet$ Pre-processing: 
We used a pre-trained 300 dimension glove vector model~\cite{pennington2014glove} to obtain the vector representation for each word.
Each utterance in the stories, question and each choice can be represented as a fixed length vector by adding the vectors of the all component words.
Before training, we pruned the utterances in the story whose vector representation has cosine distance far from the question's.
The percentage of the pruned utterances was determined by the performance of the model on the development set.
The vector representations of utterances, questions and choices were only used in this pre-processing stage and the baseline approaches in Section~\ref{sec:baseline}, not used in the proposed model.
  
$\bullet$ Training Details: 
The size of the hidden layer for both the forward and backward GRU networks were 128. 
All the bidirectional GRU networks in the proposed model shared the same set of parameters to avoid overfitting.
We used RmsProp~\cite{tieleman2012lecture} with initial learning rate of 1e-5 with momentum 0.9. 
Dropout rate was 0.2. 
Batch size was 40. 
The number of hop was tuned from 1 to 3 by development set.

\subsection{Baselines} \label{sec:baseline}
We compared the proposed model with some commonly used simple baselines in~\cite{tapaswi2015movieqa} and the memory network\cite{sukhbaatar2015end}. 

$\bullet$ Choice Length:
The most naive baseline is to select the choices based on the number of words in it without listening to the stories and looking at the questions.
This included: (i) selecting the longest choice, (ii) selecting the shortest choice or (iii) selecting the choice with the length most different from the rest choices. 
    
$\bullet$ Within-Choices similarity:
With the vector representations for the choices in pre-processing of Section~\ref{sec:setup}, we computed the cosine distance among the four choices and selected the one which is (i) the most similar to or (ii) the most different from the others.
  
$\bullet$ Question and Choice Similarity:
With the vector representations for the choices and questions in pre-processing of Section~\ref{sec:setup}, the choice with the highest cosine similarity to the question is selected. 
  
$\bullet$ Sliding Window~\cite{tapaswi2015movieqa,datar2002maintaining}:
This model try to found a window of $W$ utterances in the story with the maximum similarity to the question. 
The similarity between a window of utterances and a question was the averaged cosine similarity of the utterances in the window and the question by their glove vector representation.
After obtaining the window with the largest cosine similarity to the question, the confidence score of each choice is the average cosine similarity between the utterances in the window and the choice.
The choice with the highest score is selected as the answer.
  
$\bullet$ Memory Network~\cite{sukhbaatar2015end}:
We implemented the memory network with some modifications for this task to find out if memory network was able to deal it. 
The original memory network didn't have the embedding module for the choices, so we used the module for question in the memory network to embed the choices.
Besides, in order to have the memory network select the answer out of four choices, instead of outputting a word in its original version, we computed the cosine similarity between the the output of the last hop and the choices to select the closest choice as the answer.  
We shared all the parameters of embedding layers in the memory network for avoiding overfitting. 
Without this modification, very poor results were obtained on the testing set. 
The embedding size of the memory network was set 128, stochastic gradient descent was used as~\cite{sukhbaatar2015end} with initial learning rate of 0.01. 
Batch size was 40.
The size of hop was tuned from 1 to 3 by development set.
  

\begin{table}
\centering
\caption{Accuracy results of different models}
\label{tab:res}
\begin{tabular}{|l|l|cc|}
\hline
\multicolumn{2}{|c|}{Model}   & \multicolumn{1}{|c|}{Manual} & \multicolumn{1}{|c|}{ASR} \\ \hline
              & longest    & \multicolumn{2}{c|}{22.95\%}                             \\
(a) Choice length & shortest   & \multicolumn{2}{c|}{35.25\%}                             \\
                & different  & \multicolumn{2}{c|}{30.33\%}                             \\ \hline
(b) Within choices & similar   & \multicolumn{2}{c|}{36.07\%}                             \\
               & different & \multicolumn{2}{c|}{27.87\%}                             \\
\hline
\multicolumn{2}{|l|}{(c) Question choices}  & \multicolumn{2}{c|}{24.59\%}                             \\ \hline
\multicolumn{2}{|l|}{(d) Sliding Window}           &   33.61\% &       31.15\%             \\ \hline
\multicolumn{2}{|l|}{(e) Memory Network}           &    39.17\% &   39.17\%                 \\ \hline
(f) Our model &word           &   49.16\% &      \textbf{48.33}\%                \\
           &sentence       &   \textbf{51.67}\% &     46.67\%                  \\ \hline
\end{tabular}
\end{table}
  
  \subsection{Results}
We used the accuracy (number of question answered correctly / total number of questions) as our evaluation metric. 
The results are showed in Table~\ref{tab:res}.
We trained the model on the manual transcriptions of the stories, while tested the model on the testing set with both manual transcriptions (column labelled ``Manual'') and ASR transcriptions (column labelled ``ASR'').  
  
  $\bullet$ Choice Length: Part (a) shows the performance of three models for selecting the answer with the longest, shortest or most different length, ranging from 23\% to 35\%.
  
  $\bullet$ Within Choices similarity: Part (b) shows the performance of two models for selecting the choice which is most similar to or the most different from the others.
The accuracy are 36.09\% and 27.87\% respectively.
 
$\bullet$ Question and Choice Similarity:
In part (c), selecting the choice which is the most similar to the question only yielded 24.59\%, very close to randomly guess.
  
  $\bullet$ Sliding Window: Part (d) for sliding window is the first baseline model considering the transcription of the stories. We tried the window size \{1,2,3,5,10,15,20,30\} and found the best window size to be 5 on the development set. This implied the useful information for answering the questions is probably within 5 sentences. The performance of 31.15\% and 33.61\% with and without ASR errors respectively tells how ASR errors affected the results, and the task here is too difficult for this approach to get good results.
  
  $\bullet$ Memory Network: The results of memory network in part (e) shows this task is relatively difficult for it, even though memory network was successful in some other tasks.
  However, the performance of 39.17\% accuracy was clearly better than all approaches mentioned above, and it's interesting that this result was independent of the ASR errors and the reason is under investigation. 
  The performance was 31\% accuracy when we didn't use the shared embedding layer in the memory network.
  
  $\bullet$ \textbf{AMRNN model}: The results of the proposed model are listed in part (f), respectively for the attention mechanism on word-level and sentence-level.
  Without the ASR errors, the proposed model with sentence-level attention gave an accuracy as high as 51.67\%, and slightly lower for word-level attention.
  It's interesting that without ASR errors, sentence-level attention is about 2.5\% higher than word-level attention.
  Very possibly because that getting the information from the whole sentence is more useful than listening carefully at every words, especially for the conceptual and high-level questions in this task.
  Paying too much attention to every single word may be a bit noisy.
  On the other hand, the 34.32\% ASR errors affected the model on sentence-level more than on word-level.
  This is very possibly because the incorrectly recognized words may seriously change the meaning of the whole sentences.
  However, with attention on word-level, when a word is incorrectly recognized, the model may be able to pay attention on other correctly recognized words to compensate for ASR errors and still come up with correct answer.


  \subsection{Analysis on a typical example}
  Fig~\ref{fig:attention_example} shows the visualization of the attention weights obtained for a typical example story in the testing set, with the proposed AMRNN model using word-level or sentence-level attention on manual or ASR transcriptions respectively. 
  The darker the color, the higher the weights.
  Only a small part of the story is shown where the response of the model made good difference.
  This story was mainly talking about the thick cloud and some mysteries on Venus.
  The question for this story is \textit{``What is a possible origin of Venus'clouds?"} and the correct choice is \textit{``Gases released as a result of volcanic activity"}.
  In the manual transcriptions cases (left half of Fig~\ref{fig:attention_example}), both models, with word-level or sentence-level attention, answered the question right and focused on the core and informative words/sentences to the question.
  The sentence-level model successfully captured the sentence including \textit{``...volcanic eruptions often omits gases.''}; while the word-level model captured some important key words like \textit{``volcanic eruptions"}, \textit{``emit gases"}.
  However, in ASR cases (right half of Fig~\ref{fig:attention_example}), the ASR errors misled both models to put some attention on some irrelevant words/sentences.
  The sentence-level model focus on the irrelevant sentence \textit{``In other area, you got canyons..."}; while the word-level model focused on some irrelevant words \textit{``canyons"}, \textit{``rift malaise"}, but still capture some correct important words like \textit{``volcanic"} or \textit{``eruptions"} to answer correctly.
  By the darkness of the color, we can observe that the problem caused by ASR errors was more serious for the sentence-level attention when capturing the key concepts needed for the question.
  This may explain why in part (f) of Table~\ref{tab:res} we find degradation caused by ASR errors was less for word-level model than for sentence-level model.
  
  \section{Conclusions} \label{conclu}
In this paper we create a new task with the TOEFL corpus.
TOEFL is an English examination, where the English learner is asked to listen to a story up to 5 minutes and then answer some corresponding questions.
The learner needs to do deduction, logic and summarization for answering the question.
We built a model which is able to deal with this challenging task.
On manual transcriptions, the proposed model achieved 51.56\% accuracy, while the very capable memory network got only 39.17\% accuracy. 
Even on ASR transcriptions with WER of 34.32\%, the proposed model still yielded \textbf{48.33}\% accuracy.
We also found that although sentence-level attention achieved the best results on the manual transcription, word-level attention outperformed the sentence-level when there were ASR errors.
    
  \newpage
  \eightpt
  \bibliographystyle{IEEEtran}
  \bibliography{mybib,Lee}


\end{document}